\documentclass[runningheads]{llncs}
\usepackage[T1]{fontenc}

\renewcommand{\refname}{References}

\usepackage{cite}
\usepackage{amsmath,amssymb,amsfonts}
\usepackage{algorithmic}
\usepackage{graphicx}
\usepackage{textcomp}
\usepackage{xcolor}
\usepackage{tabularx}
\usepackage{hyperref}
\hypersetup{
    colorlinks=true, 
    linkcolor=black,   
    citecolor=black,   
    urlcolor=black,    
}

\usepackage{mathptmx}

\usepackage[utf8]{inputenc}
\usepackage[T1]{fontenc}
\usepackage{multirow}
\usepackage{pifont}
\usepackage{booktabs}
\usepackage{array}
\usepackage{makecell}

\makeatletter
\let\NAT@parse\undefined
\makeatother
\usepackage{hyperref}
\usepackage{amsfonts} 
\usepackage{amssymb} 
\usepackage{graphicx}
\usepackage{makecell}
\usepackage{amsmath}
\usepackage{CJK}
\usepackage{array} 
\usepackage{graphicx} 
\RequirePackage{CJKnumb}
\usepackage{caption}
\usepackage{graphicx, subfig}
\usepackage{amsmath} 

\usepackage{graphicx}
\usepackage{marvosym} 

\begin{document}
\title{Landmark-Guided Knowledge for Vision-and-Language Navigation }

\titlerunning{Landmark-Guided VLN}
%





\author{
   Dongsheng Yang\inst{1} \and 
   Meiling Zhu\inst{2}\textsuperscript{(\Letter)} \and 
   Yinfeng Yu\inst{1}\textsuperscript{(\Letter)}
 }



 \institute{
 School of Computer Science and Technology, Xinjiang University, Urumqi, China \\
 \email{yuyinfeng@xju.edu.cn}
 \and
 No. 59 Middle School of Urumqi, Urumqi, China \\
 \email{514472795@qq.com}
 }

\renewcommand{\thefootnote}{}  

%
%

\maketitle              
\begin{abstract}
Vision-and-language navigation is one of the core tasks in embodied intelligence, requiring an agent to autonomously navigate in an unfamiliar environment based on natural language instructions. However, existing methods often fail to match instructions with environmental information in complex scenarios, one reason being the lack of common-sense reasoning ability. This paper proposes a vision-and-language navigation method called Landmark-Guided Knowledge (LGK), which introduces an external knowledge base to assist navigation, addressing the misjudgment issues caused by insufficient common sense in traditional methods. Specifically, we first construct a knowledge base containing 630,000 language descriptions and use knowledge Matching  to align environmental subviews with the knowledge base, extracting relevant descriptive knowledge. Next, we design a Knowledge-Guided by Landmark (KGL) mechanism, which guides the agent to focus on the most relevant parts of the knowledge by leveraging landmark information in the instructions, thereby reducing the data bias that may arise from incorporating external knowledge. Finally, we propose Knowledge-Guided Dynamic Augmentation (KGDA), which effectively integrates language, knowledge, vision, and historical information. Experimental results demonstrate that the LGK method outperforms existing state-of-the-art methods on the R2R and REVERIE vision-and-language navigation datasets, particularly in terms of navigation error, success rate, and path efficiency.

\end{abstract}
\section{Introduction}

Vision-and-language navigation (VLN) \cite{anderson2018vision, qi2020reverie, qiao2022hop, an2024etpnav} is one of the core tasks in embodied artificial intelligence (embodied AI)\cite{YinfengIJCAI2023MACMA,SAAVN}, which requires an agent to navigate in an unseen environment by following natural language instructions. This task poses significant challenges as it not only requires precise parsing of language instructions to extract implicit goals and navigation cues, but also necessitates accurate visual perception to construct a semantic understanding of the surrounding environment. To address these challenges, recent research has focused on developing deep models capable of enabling effective cross-modal reasoning between language and vision.

Thanks to the widespread use of transformers \cite{vaswani2017attention}, methods now primarily adopt transformer frameworks to build agent models \cite{lin2022adapt, wang2024lookahead, gao2023adaptive}, differing from earlier approaches that used recurrent neural networks inefficiently for learning long-term dependencies in agent modeling \cite{anderson2018vision, wang2019reinforced}. The self-attention mechanism of transformers allows for better capturing long-term dependencies across both modalities, resulting in more effective spatiotemporal modeling.

In this study, we focus on the indoor VLN task defined in discrete environments, which includes both coarse-grained goal-directed navigation (e.g., REVERIE~\cite{qi2020reverie}) and fine-grained step-by-step instruction following (e.g., R2R~\cite{anderson2018vision}). Despite the impressive results achieved by existing methods on these benchmarks \cite{he2024frequency, lin2025navcot, cui2023grounded}, several challenges persist in real-world scenarios.

\begin{figure*}[ht!]  
    \centering  
    \includegraphics[width=\textwidth]{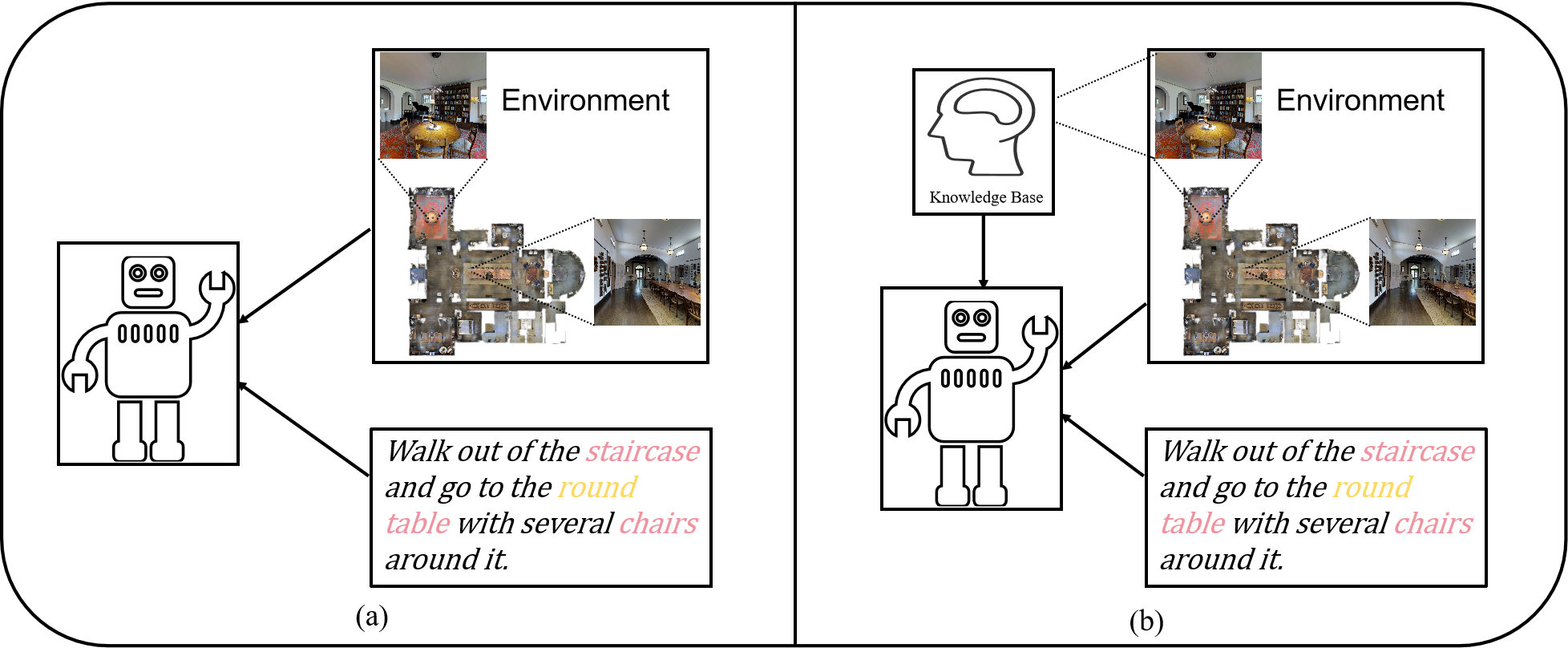}  
    \caption{The difference between the LGK method and other approaches. When the agent is confused by identical objects with different attributes, LGK assists navigation by introducing knowledge.}  
    \label{fig:duibitu}  
\end{figure*}

As shown in Figure~\ref{fig:duibitu}(a), most VLN models \cite{wang2023gridmm, zhou2025navgpt, chen2022think} rely only on visual observations and instruction cues as inputs. However, in complex environments with multiple candidate paths and visually similar landmarks (e.g., numerous chairs and tables), an agent may struggle to resolve subtle semantic ambiguities—such as distinguishing between different types of tables—without external knowledge or common-sense priors.

As shown in Figure~\ref{fig:duibitu}(b), we propose introducing a large-scale language-based knowledge base into the navigation process to address this issue. By leveraging a pre-constructed knowledge base containing 630,000 descriptive sentences, the agent can retrieve additional semantic cues to supplement visual observations. We use CLIP \cite{radford2021learning} to align subview images with relevant knowledge fragments, thereby enriching the agent’s understanding of the environment.

Building on DUET~\cite{chen2022think}, we introduce a novel framework called Landmark-Guided Knowledge (LGK) that provides the agent with descriptive common-sense knowledge based on visual landmarks. The LGK framework consists of three core modules: (1) Knowledge Matching, which retrieves semantically relevant descriptions from the knowledge base; (2) Knowledge-Guided by Landmark, which guides knowledge selection based on landmarks mentioned in the instruction, thereby reducing data bias introduced by external knowledge and allowing the model to focus more effectively on landmark-relevant information; and (3) Knowledge-Guided Dynamic Augmentation, which enhances navigation reasoning in two ways: on the one hand, by augmenting the instruction with external knowledge and dynamically balancing the enhanced instruction with the original one; on the other hand, by strengthening the relationship between the local environment and the knowledge-interacted objects through augmentation and adjustment.

Experimental results on the R2R and REVERIE datasets demonstrate that the LGK method outperforms existing state-of-the-art approaches across multiple evaluation metrics, validating the effectiveness of landmark-guided knowledge enhancement in VLN tasks. The main contributions of this paper are as follows:

(1) We propose a novel Landmark-Guided Knowledge (LGK) method, which effectively addresses semantic ambiguity caused by insufficient common sense in traditional vision-and-language navigation methods by introducing a large-scale language description knowledge base and employing knowledge Matching (KM).

(2)  We design two modules: Knowledge-Guided by Landmark (KGL) and Knowledge-Guided Dynamic Augmentation (KGDA). The former utilizes key landmarks in the instructions to fine-tune the retrieved knowledge, effectively reducing semantic noise and data bias. The latter achieves the enhancement and dynamic fusion of the language, vision, and knowledge modalities, enabling the model to more accurately focus on key semantic information among multiple candidate targets.

(3) Extensive experiments on the R2R and REVERIE datasets show that the LGK method outperforms existing state-of-the-art methods in multiple metrics, such as navigation error, success rate, and path efficiency, demonstrating the method's effectiveness and generality in complex scenarios.

\section{Related Work}
\subsection{Vision-and-Language Navigation}
Vision-and-language navigation (VLN) \cite{anderson2018vision, qi2020reverie, qiao2022hop, an2024etpnav} is one of the core tasks in embodied artificial intelligence (embodied AI), aiming to enable agents to autonomously navigate in unknown environments by following natural language instructions. This task is challenging as it requires the agent to accurately parse language instructions to extract implicit goals and navigation cues and effectively perceive visual information to build a semantic understanding of the surrounding environment. To tackle these challenges, recent research has focused on developing deep learning models capable of effectively achieving cross-modal reasoning between language and vision.

Early VLN methods primarily relied on recurrent neural networks (RNNs) and long short-term memory networks (LSTMs) to jointly encode visual and textual information \cite{anderson2018vision, fried2018speaker, tan2019learning}. However, these methods often struggled to capture long-term contextual information when processing long trajectories. As research progressed, transformer-based VLN methods emerged, enabling more effective joint representations of visual and textual information, which improved overall model performance \cite{chen2021history, chen2022think, hong2021vln, schumann2024velma}.

To better capture long-term dependencies between historical observations and actions, HAMT \cite{chen2021history} proposed using transformers directly to model the relationships between historical information and actions. However, this method only supports local actions and does not effectively address global planning problems. To solve this issue, DUET \cite{chen2022think} introduced a graph transformer, which combines local observations and global topological maps to facilitate action planning and global behavior generation. BEVBert \cite{an2023bevbert} brought a hybrid topological-metric map into VLN, where the topological map is used for long-term planning, while the metric map aids in short-term reasoning.

With the significant progress of large language models (LLMs) in natural language processing and instruction-following tasks, NavGPT2 \cite{zhou2024navgpt} proposed applying LLMs to VLN tasks to enhance the model’s knowledge reasoning ability. VELMA \cite{schumann2024velma} integrated LLMs and used trajectory and visual environment observations as contextual cues for the next action, further advancing the development of VLN tasks.

\subsection{Knowledge Bases in Vision-and-Language Navigation}

In recent years, more studies have focused on integrating external knowledge bases into vision-and-language navigation (VLN) to enhance the model's understanding and reasoning capabilities. Existing large-scale structured knowledge bases, such as ConceptNet \cite{liu2004conceptnet} and DBpedia \cite{auer2007dbpedia}, provide rich common-sense knowledge through automated data extraction and manual annotation. CKR \cite{gao2021room} utilizes common-sense information from ConceptNet to learn the relationships between room entities and objects. KERM \cite{li2023kerm} constructs an external knowledge base from Visual Genome \cite{krishna2017visual} to assist in establishing relationships between various entities in the instruction. ACK \cite{mohammadi2024augmented} builds a knowledge base by retrieving common-sense information from ConceptNet and uses a refinement module to remove noise and irrelevant knowledge.

In contrast to these works, this study focuses on guiding the knowledge matched from the knowledge base by emphasizing important information in navigation instructions, particularly landmark entities. This approach helps direct the focus toward task-relevant key elements, thereby improving the reasoning and decision-making abilities of the navigation agent in complex environments.

\section{Method}
The LGK framework is set in a discrete environment constructed by the Matterport3D simulator \cite{anderson2018vision}, which is represented by an undirected graph \( G = \{V, E\} \), where the set of nodes \( V = \{V_i\}_{i=1}^{K} \) represents all navigable nodes (with a total of \( K \) nodes), and the set of edges \( E \) describes the connections between these nodes. In the R2R \cite{anderson2018vision} and REVERIE \cite{qi2020reverie} datasets, the agent is placed at a predefined location at the beginning of the task, equipped with a GPS sensor and an RGB camera. Its goal is to autonomously explore an unseen environment and eventually navigate to the target location based on natural language instructions. The natural language instruction is represented as \( I = \{w_i\}_{i=1}^{L} \), where \( L \) is the number of words in the instruction, and each word has been embedded into a vector space.

At each time step \( t \), the agent updates its position based on the GPS sensor and captures a panoramic view with the RGB camera. The panoramic view consists of 36 individual images, denoted as \( R_t = \{r_i\}_{i=1}^{36} \), where each image \( r_i \) is described by its feature vector and its unique orientation \( (\theta_{1, i}, \theta_{2, i}) \), where \( \theta_{1, i} \) and \( \theta_{2, i} \) represent the azimuth and elevation angles, respectively. For target localization tasks, \( N \) object features \( O_t = \{o_i\}_{i=1}^{N} \) are also extracted from the panoramic view using annotated bounding boxes or an object detector.

Next, a multi-layer Transformer \cite{vaswani2017attention} is employed to model the spatial relationships between the images and objects, represented as \( [R_t, O_t] \).

During the navigation process, the agent selects actions from the action space based on the current node's panoramic view and the natural language instruction, aiming to move to the next navigable node until the agent reaches the target location or the maximum number of steps is reached. Once a stop action is issued, the agent predicts the target object's location. If the distance between the target location and the agent's position is less than or equal to 3 meters, the task is considered successfully completed.
\subsection{Method Overview}
\label{subsec:overview}

\begin{figure*}[ht!]
    \centering
    \includegraphics[width=\textwidth]{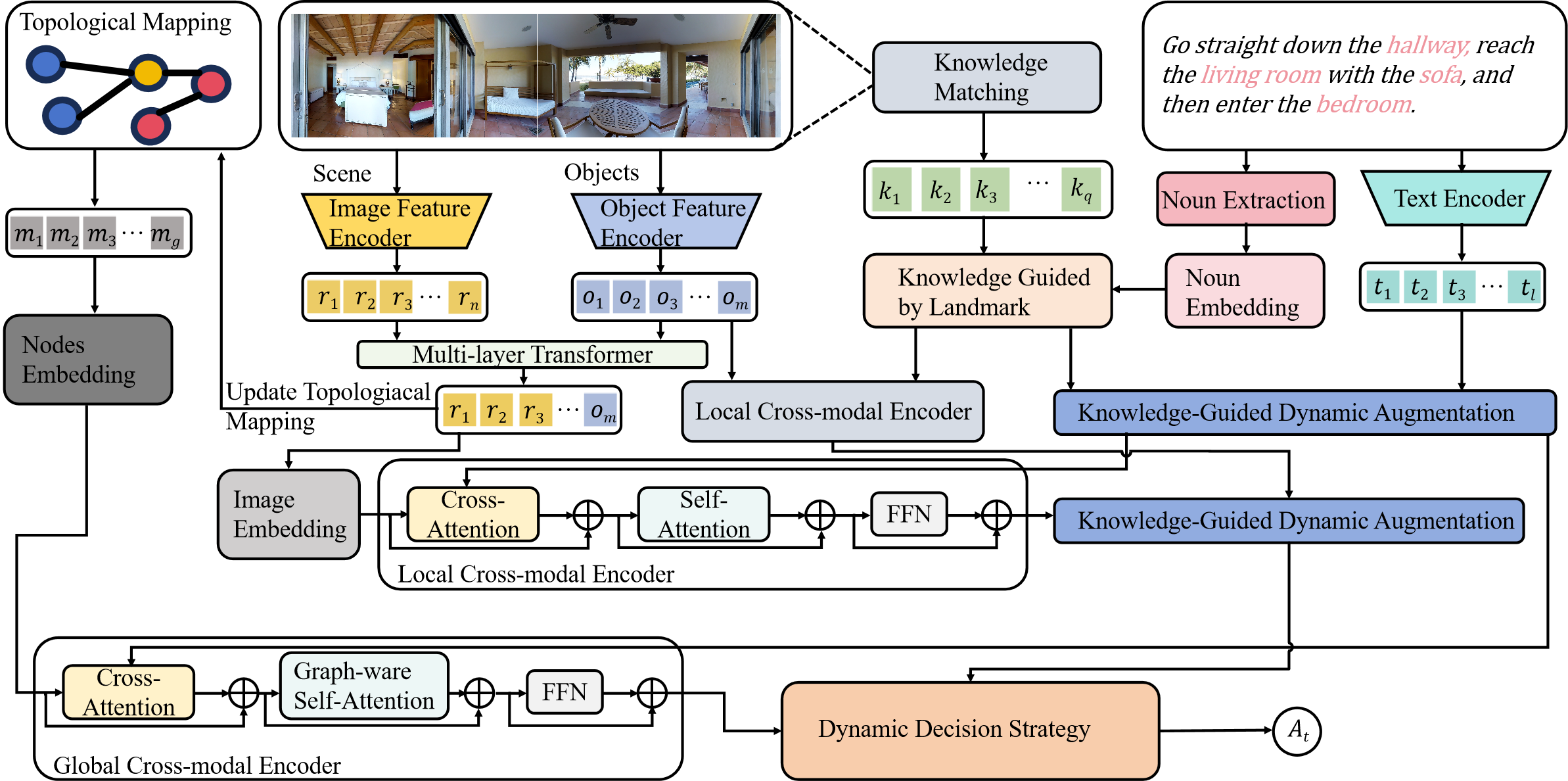} 
    \caption{Overview of the proposed Landmark-Guided Knowledge Network (LGK) structure. The network incorporates an external knowledge base to assist the agent in navigation, focusing on three key components designed around the knowledge base: (1) Knowledge Matching, (2) Knowledge-Guided by Landmark, and (3) Knowledge-Guided Dynamic Augmentation.}
    \label{fig:kjt}
\end{figure*}

As shown in Figure \ref{fig:kjt}, compared to the baseline DUET model \cite{chen2022think}, our model not only takes visual information, natural language instructions, and a dynamically updated topological map as inputs, but also incorporates an external knowledge base to enhance the model's environmental understanding and navigation capability. To better utilize the knowledge base, we build a series of modules around it, aiming to enable the agent to fully leverage knowledge to assist reasoning. In this section, we will sequentially introduce the process of constructing the knowledge base, the Knowledge Matching, the Knowledge Guided by Landmark, and the Knowledge-Guided Dynamic Augmentation.

\subsection{Knowledge Acquisition}
\label{subsec:tse}

In vision-and-language navigation, when the agent encounters objects of the same type but with different attributes (e.g., a rectangular table vs. a circular table), errors in decision-making often occur due to a lack of common-sense reasoning. To minimize the mistakes caused by common-sense issues, we design a knowledge base to assist the agent in navigation.

\subsubsection{Knowledge Base Construction}
\label{subsubsec:oap}

The knowledge base serves as a further description of the environment, aiming to provide a detailed textual representation of the environment. We follow the method of KERM \cite{li2023kerm}, constructing the knowledge base from the Visual Genome dataset \cite{krishna2017visual}. The attribute annotations in this dataset are in the form of "attribute-object" pairs, and we convert all "attribute-object" pairs and "subject-verb-object" triples into their synonymous canonical forms, resulting in a knowledge base containing 630,000 language descriptions (\(KB_1\)).
\label{subsec:topa}
\begin{figure}[ht]
    \centering
    \includegraphics[width=3.35in]{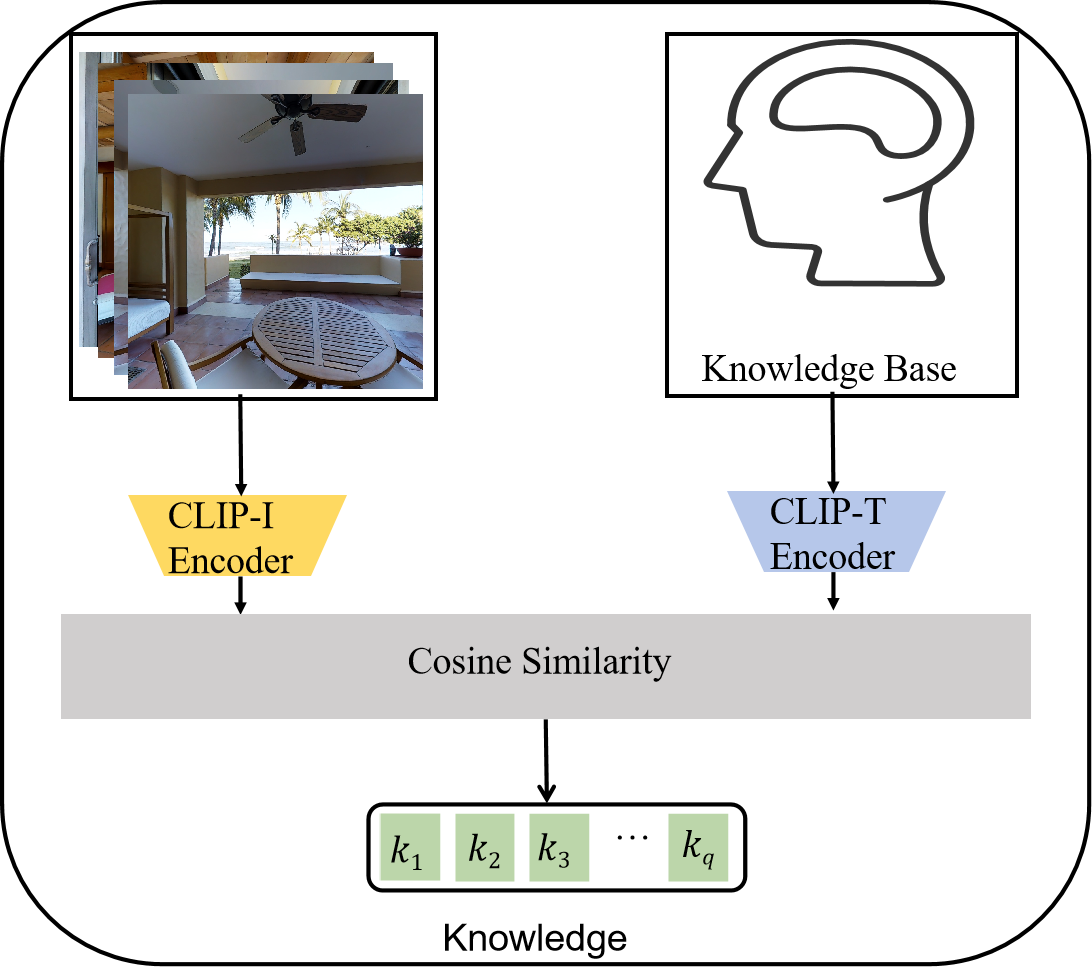}
    \caption{Overview of Knowledge Matching, using CLIP to encode both environmental and knowledge information, followed by cosine similarity calculation.}
    \label{fig:km}
\end{figure}

\subsubsection{Knowledge Matching}

During navigation, our goal is to enable the agent to match relevant knowledge from the knowledge base in real-time based on the current environment. To achieve this, we propose the Knowledge Matching, as shown in Figure \ref{fig:km}. To improve the accuracy of matching environmental information, we divide each single view into 5 sub-regions and process the knowledge base (\(KB_1\)) further. We use Spacy's language model (\texttt{en\_core\_web\_sm}) for noun extraction from the knowledge base, and for phrases without nouns, we replace them with "None," resulting in a new knowledge base of 630,000 entries that only contains nouns (\(KB_2\)). The knowledge positions in \(KB_2\) correspond one-to-one with those in \(KB_1\). Then, we use the pre-trained CLIP model \cite{radford2021learning} to encode the descriptive knowledge. CLIP consists of the CLIP-I image encoder and the CLIP-T text encoder. We use CLIP-T to encode all noun information from \(KB_2\) as search keys and CLIP-I to encode the environmental information of each sub-region as queries. We then compute the nouns from \(KB_2\) that match the subview most closely, retaining the top 5 nouns with the highest cosine similarity score for each sub-region, and replace these nouns with the descriptive knowledge from \(KB_1\).

\begin{figure}[ht]
    \centering
    \includegraphics[width=3.35in]{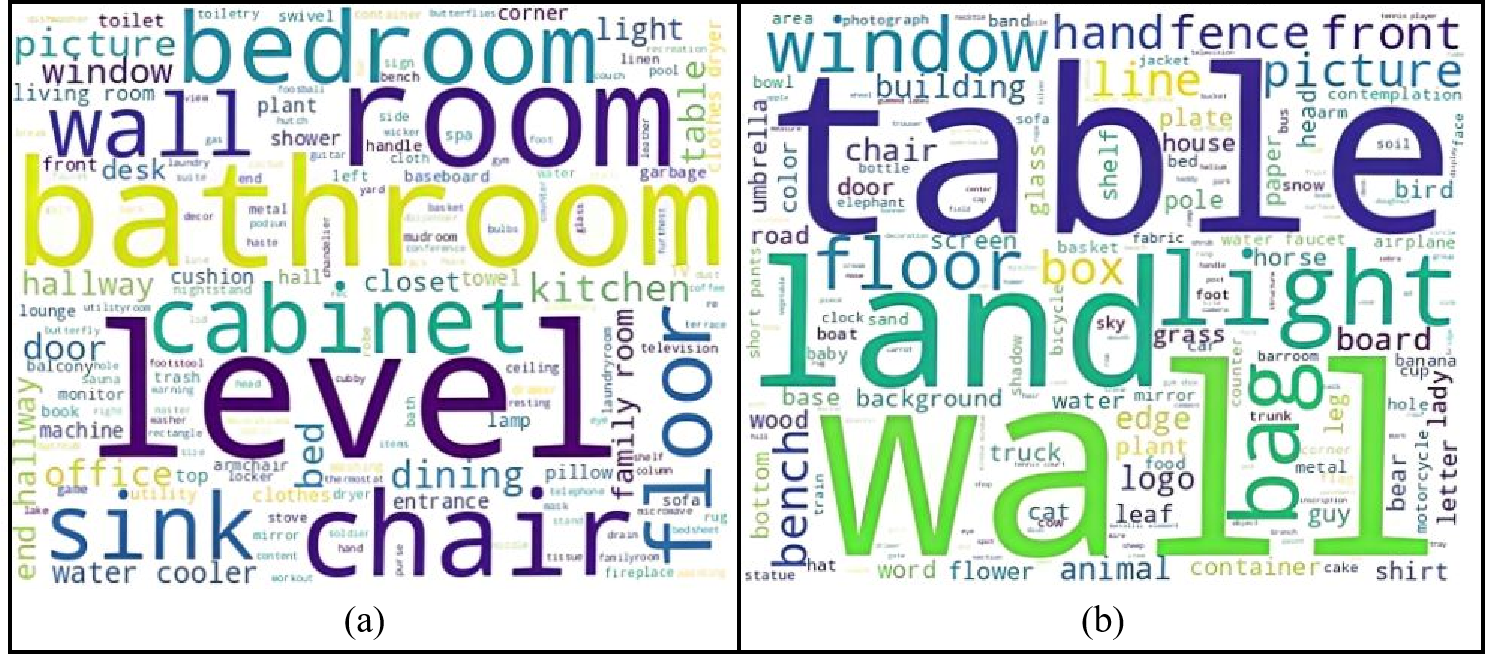}
    \caption{(a) Word cloud of the REVERIE dataset and (b) word cloud of the knowledge base.}
    \label{fig:word_clouds}
\end{figure}

In vision-and-language navigation, not all information in natural language instructions is equally essential; landmark information is particularly crucial for the agent. Therefore, we extract noun phrases from the instructions, which often contain landmark names. As shown in Figure \ref{fig:word_clouds}, (a) represents the word cloud of nouns from the REVERIE dataset, and (b) shows the word cloud of nouns from the knowledge base. While the knowledge base provides common-sense knowledge to the agent, it may also introduce data bias. To reduce the impact of data bias and enhance the agent’s perception of landmarks, we design the Knowledge Guided by Landmark(KGL).

\subsection{Knowledge Guided by Landmark}
\label{subsec:iopa}

\begin{figure}[!t]
    \centering
    \includegraphics[width=3.35in]{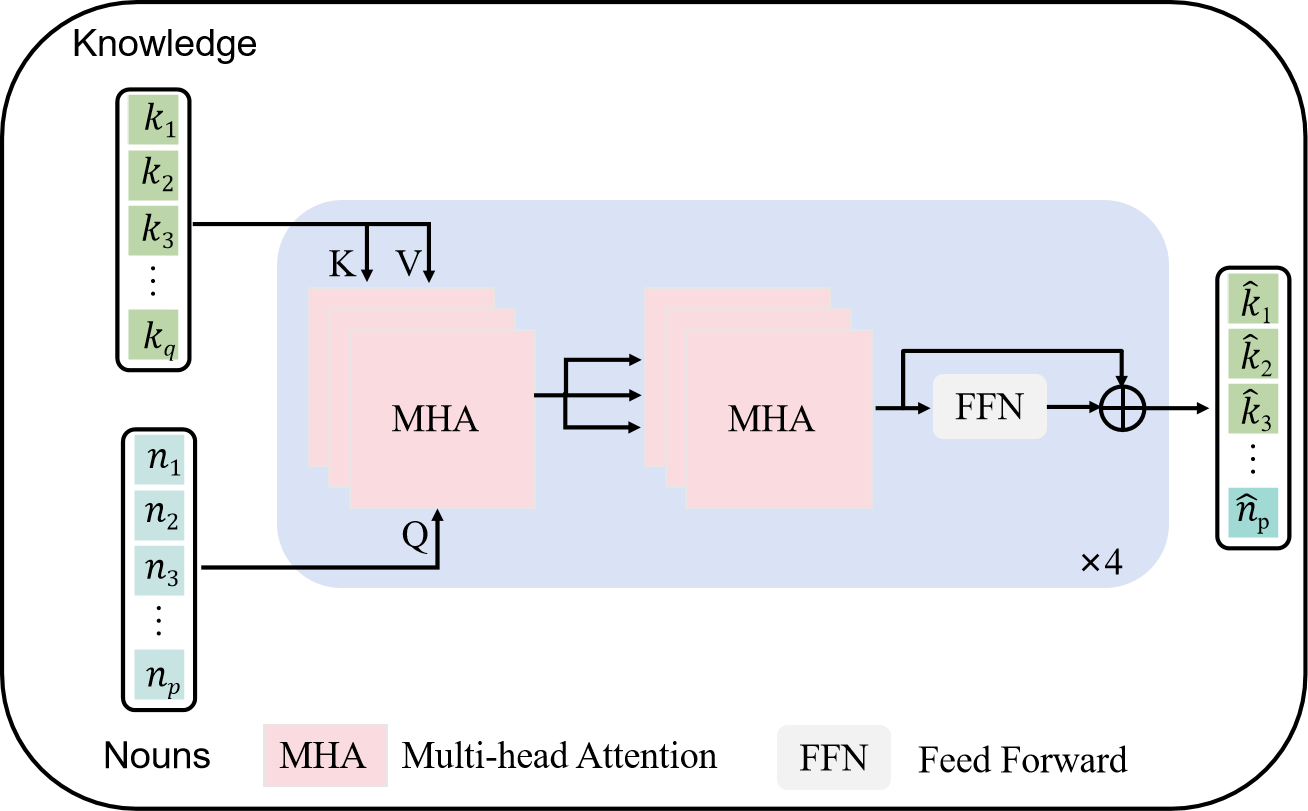}
    \caption{Knowledge Guided by Landmark.}
    \label{fig:kgl}
\end{figure}

As shown in Figure \ref{fig:kgl}, the KGL consists of a cross-attention layer, a self-attention layer, and a feedforward network. At time step \( t \), the noun phrases (i.e., landmarks) \( N = \{n_i\}_{i=1}^{p} \) (where \( p \) is the number of nouns) from the natural language instruction and the environmental knowledge \( K = \{k_i\}_{i=1}^{q} \) (where \( q \) is the number of knowledge entries) are fed into the cross-attention layer of KGL. We use the knowledge \( K \) as the query (\( q \)), and the nouns \( N \) as the key (\( k \)) and value (\( v \)), expressed as:
\begin{align}
\tilde{K}_q = \text{CrossAttn}(N'_p, K'_q) .
\end{align}
Then, \( \tilde{K}_q \) is passed into the self-attention layer, generating the landmark-guided knowledge-enhanced features \( \hat{K}_q \).

\subsection{Knowledge-Guided Dynamic Augmentation}
\label{subsubsec:object_embedding}

\subsubsection{Cross-Modal Encoder}
\label{subsubsec:cross_modal_encoder}

After obtaining the image object features \( O_t \) and landmark-guided knowledge-enhanced features \( \hat{K}_q \), we fuse these two features using a local cross-modal encoder to learn their correlation. Specifically, we employ the LXMERT model \cite{tan2019lxmert} as the cross-modal encoder, which models the relationship between image object features and knowledge-enhanced features, ultimately obtaining the fused image object features \( \overline{O}_t \).

\subsubsection{Knowledge-Guided Dynamic Augmentation}

\begin{figure}[!t]
    \centering
    \includegraphics[width=3.35in]{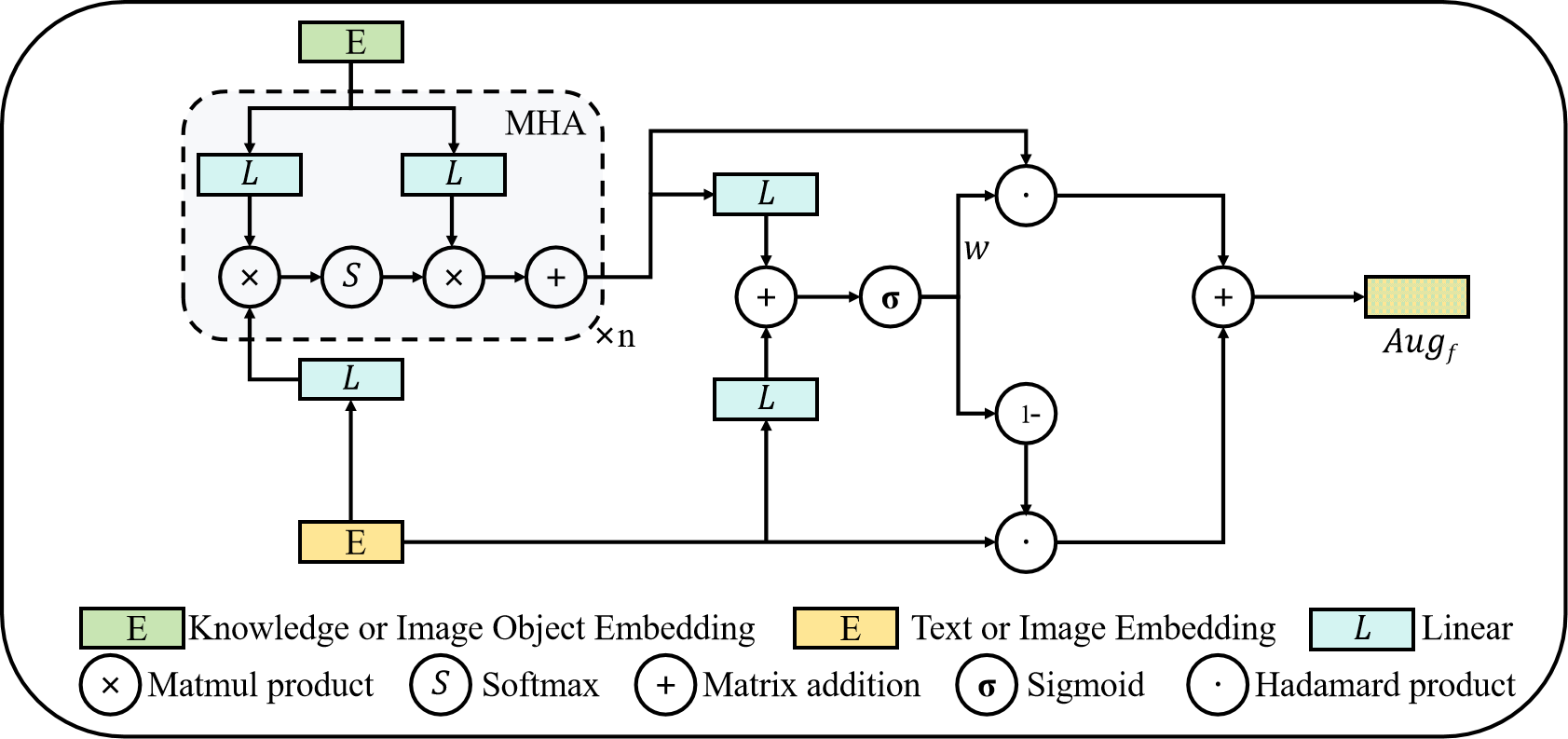}
    \caption{Knowledge-Guided Dynamic Augmentation.}
    \label{fig:kgda}
\end{figure}

After obtaining the landmark-guided knowledge-enhanced features \( \hat{K}_q \), we input them along with the instruction features \( I'_L \) into the Knowledge-Guided Dynamic Augmentation, as shown in Figure \ref{fig:kgda}. We introduce a multi-head attention mechanism \cite{vaswani2017attention}, using the knowledge features \( \hat{K}_q \) as the key and value, and the instruction features \( I'_L \) as the query to update the relationship between context features and knowledge, ultimately obtaining the knowledge-enhanced instruction features \( I''_L \). To dynamically adjust the ratio between the enhanced instruction features and the original instruction features, we adopt a residual dynamic fusion method, where the weight for \( I''_L \) is dynamically computed using a Sigmoid function, and then the enhanced instruction features \( A_f^I \) are obtained via weighted summation. The knowledge-guided dynamic instruction enhancement process can be expressed as:
\begin{align}
I''_L &= \text{MHA}(I'_L, \hat{K}_q) .\label{eq3} \\
\omega &= \delta (I'_L W_g + I''_L W_c + b_I) .\label{eq4} \\
A_f^I &= \omega \odot I''_L + (1 - \omega) \odot I'_L . \label{eq5}
\end{align}

Where \( W_g \) and \( W_c \) are learnable parameters, \( \delta \) denotes the sigmoid activation function, and \( \odot \) denotes element-wise multiplication.

After obtaining the enhanced instruction features \( A_f^I \), we use the local cross-modal encoder to model the relationship between the panoramic view features \( [R_t, O_t] \) and the instruction features \( A_f^I \), thus obtaining the local features \( [\hat{R}_t, \hat{O}_t] \). To dynamically adjust the ratio between \( \overline{O}_t \) and \( [\hat{R}_t, \hat{O}_t] \), we introduce the Knowledge-Guided Dynamic Augmentation to obtain the enhanced local features \( [\hat{R}'_t, \hat{O}'_t] \). For the topological map \( M = \{m_i\}_{i=1}^{N} \), the interaction with the instruction follows a similar process, and by calling the global encoder, we obtain the global features \( [\hat{M}_t] \).

\subsection{Dynamic Fusion Strategy}
\label{subsec:dynamic_fusion}

After processing by the local encoder and global cross-modal encoder, we use a dynamic decision module to select actions. Specifically, we first extract the [CLS] token from the topological map features \( [\hat{M}_t] \) and the panoramic view features \( [\hat{R}'_t, \hat{O}'_t] \), denoted as \( A_l \) and \( A_g \), respectively. We concatenate these and compute the scalar weights for the global and local branches using a feedforward network (FFN). Next, we use two additional FFNs to project \( [\hat{M}_t] \) and \( [\hat{R}'_t, \hat{O}'_t] \) into the score domain, converting the local action score \( A_l \) into the global action score \( \hat{A}_l \), and finally obtain the action prediction probability \( A_t \) through weighted summation. The process can be expressed as:
\begin{align}
\sigma &= \delta(\text{FFN}([A_g, A_l])) . \\
G_g &= \text{FFN}([\hat{M}_t]), \quad G_l = \text{FFN}([\hat{R}'_t, \hat{O}'_t]). \\
A_t &= \sigma G_g + (1 - \sigma) \hat{G}_l.
\end{align}

\section{Experiments}
\subsection{Datasets}
We evaluate the effectiveness of our model on two popular vision-and-language navigation datasets: R2R \cite{anderson2018vision} and REVERIE \cite{qi2020reverie}. Both datasets are built upon Matterport 3D \cite{Matterport3D}, consisting of 90 indoor scenes, which are divided into training, validation-seen (val-seen), validation-unseen (val-unseen), and test (test-unseen) sets.
The R2R dataset comprises 7,189 navigation paths built within the Matterport 3D environment. These paths contain 21,567 English step-by-step navigation instructions, with an average length of 32 words. Of these, 14,025 instructions are used for training, 1,020 for validation on val-seen, 4,173 for validation on val-unseen, and 2,349 for testing.
The REVERIE dataset primarily contains high-level navigation instructions that describe target locations, with each instruction having an average of 21 words. Each panoramic image provides predefined object bounding boxes, and at the end of the navigation, the agent must select the correct object bounding box. Each panoramic sub-image contains an average of 10 objects.

\subsection{Evaluation Metrics}
We use standard VLN evaluation metrics to assess the model's navigation performance. On the R2R dataset, we use the following metrics:
(1) Navigation Error (NE): The average distance (in meters) between the agent's stopping position and the actual target position.
(2) Success Rate (SR): The proportion of paths where the agent's stopping position is within 3 meters of the actual target position.
(3) Oracle Success Rate (OSR): The success rate when using the Oracle stopping strategy.
(4) Success Weighted by Path Length (SPL): A metric that combines success rate and path efficiency, penalizing path length.
On the REVERIE dataset, in addition to the above metrics, we also use the following metrics:
(1) Remote Goal Success Rate (RGS): The proportion of correct object localization at the target position.
(2) Remote Goal Success Rate Weighted by Path Length (RGSPL): A metric that combines RGS and path length.
\begin{table*}[!h]
\centering
\caption{Comparison with the state-of-the-art methods on the R2R dataset}
\label{tab:r2r}
\resizebox{0.97\textwidth}{!}{%
\begin{tabular}{lccccccccccccc}
\toprule
Methods & \multicolumn{4}{c}{Val Seen} & \multicolumn{4}{c}{Val Unseen} & \multicolumn{4}{c}{Test Unseen} \\
\cmidrule(lr){2-5} \cmidrule(lr){6-9} \cmidrule(lr){10-13}
 & NE$\downarrow$ & OSR$\uparrow$ & SR$\uparrow$ & SPL$\uparrow$
 & NE$\downarrow$ & OSR$\uparrow$ & SR$\uparrow$ & SPL$\uparrow$
 & NE$\downarrow$ & OSR$\uparrow$ & SR$\uparrow$ & SPL$\uparrow$ \\ 
\midrule
Seq2Seq\cite{anderson2018vision} & 6.01 & - & 39 & - & 7.81 & - & 22 & - & 7.85 & - & 20 & 18 \\ 
EnvDrop \cite{tan2019learning} & 3.99 & - & 62 & 59 & 5.22 & - & 52 & 48 & 5.23 & 59 & 51 & 47 \\ 
EnvEdit \cite{li2022envedit} & 2.17 & - & 77 & 74 & 3.24 & - & 69 & 64 & 3.59 & - & 68 & 64 \\  
HOP+ \cite{qiao2023hop+} & 2.33 & - & 78 & 73 & 3.49 & - & 67 & 61 & 3.71 & - & 66 & 60 \\ 
HAMT \cite{chen2021history} & 2.34 & 82 & 76 & 72 & 2.29 & 73 & 66 & 61 & 3.93 & 72 & 65 & 60 \\ 
DUET \cite{chen2022think} & 2.28 & 86 & 79 & 73 & 3.31 & 81 & 72 & 60 & 3.65 & 76 & 69 & 59 \\ 
KERM \cite{li2023kerm} & 2.19 & - & 80 & 74 & 3.22 & - & 72 & 61 & 3.61 & - & 70 & 59 \\ 
GridMM \cite{wang2023gridmm} & 2.34 & - & 80 & 74 & 2.83 & - & \textbf{75} & 64 & 3.35 & - & 73 & 62 \\ 
LSAL \cite{wu2024vision} & 2.88 & - & 73 & 70 & 3.62 & - & 65 & 59 & 4.00 & - & 63 & 58 \\ 
NavGPT2 \cite{zhou2024navgpt} & 2.84 & 83 & 74 & 63 & 2.98 & \textbf{84}  & 74 & 61 & 3.33 & 80 & 72 & 60 \\ 
KESU \cite{gao2024enhancing} & 2.19 & - & 81& 75 & 2.96 & - & 73 & 62 & 3.31 & - & 72 & 61 \\
BEVBert \cite{an2023bevbert} & 2.17 & - & 81 & 74 & 2.81 & - & \textbf{75} & 64 & 3.13 & \textbf{81} & 73 & 62 \\ 
\textbf{LGK(Ours)} & \textbf{2.10} & \textbf{88} & \textbf{82} & \textbf{76} & \textbf{2.69} & \textbf{84} &  \textbf{75} & \textbf{65} & \textbf{3.09} & \textbf{81} & \textbf{74} & \textbf{64} \\
\bottomrule
\end{tabular}
}
\end{table*}

\begin{table*}[h]
\centering
\caption{Comparison with the state-of-the-art methods on the REVERIE dataset}
\label{tab:reverie}
\resizebox{0.97\textwidth}{!}{%
\begin{tabular}{lcccccccccc}
\toprule
Methods& \multicolumn{5}{c}{Val seen} & \multicolumn{5}{c}{Test Unseen} \\
\cmidrule(lr){2-6} \cmidrule(lr){7-11}
& OSR$\uparrow$ & SR$\uparrow$ & SPL$\uparrow$ & RGS$\uparrow$ & RGSPL$\uparrow$ 
& OSR$\uparrow$ & SR$\uparrow$ & SPL$\uparrow$ & RGS$\uparrow$ & RGSPL$\uparrow$ \\
\midrule
Airbert\cite{guhur2021airbert} & 48.98 & 47.01 & 42.34 & 32.75 & 30.01 & 34.20 & 30.28 & 23.61 & 16.83 & 13.28 \\ 
HOP+\cite{qiao2023hop+} & 56.43 & 55.87 & 49.55 & 40.76 & 36.22 & 35.81 & 33.82 & 28.24 & 20.20 & 16.86 \\ 
HAMT\cite{chen2021history} & 47.65 & 43.29 & 40.19 & 27.20 & 25.18 & 33.41 & 30.40 & 26.67 & 14.88 & 13.08 \\ 
DUET\cite{chen2022think} & 73.86 & 71.75 & 63.74 & 57.41 & 51.14 & 56.91 & 52.51 & 36.06 & 31.88 & 22.06 \\ 
LANA \cite{wang2023lana} & 74.28 & 71.94 & 62.77 & 59.02 & 50.34 & 57.20 & 51.72 & 36.45 & 32.95 & 22.85 \\ 
KERM\cite{li2023kerm} & 74.49 & 71.89 & 64.04 & 57.55 & 51.22 & 57.44 & 52.26 & 37.46 & 32.69 & 23.15 \\ 
CONSOLE\cite{lin2024correctable} & 76.25 & 74.14 & 65.15 & 60.08 & 52.69 & 59.60 & 55.13 & 37.13 & 33.18 & 22.25 \\ 
ACME\cite{wu2025adaptive} & 74.94 & 71.96 & 64.45 & 57.20 & 51.28 & 57.48 & 51.89 & 34.65 & 33.12 & 23.57 \\ 
BEVBert\cite{an2023bevbert} & 76.18 & 73.72 & 65.32 & 57.70 & 51.73 & 57.26 & 52.81 & 36.41 & 32.06 & 22.09 \\ 
VER\cite{liu2024volumetric} & \textbf{80.09} & 75.83 &66.19& 61.71 & 56.20 & 62.22 & 56.82 & 38.76 & 33.88 & 23.19 \\ 
\textbf{LGK(Ours)} & 79.83 & \textbf{77.37} & \textbf{69.49} & \textbf{62.54} & \textbf{56.32} & \textbf{62.96} & \textbf{57.68} & \textbf{40.63}  & \textbf{35.28} & \textbf{24.91} \\
\bottomrule
\end{tabular}
}
\end{table*}

\subsection{Implementation Details}
Our model consists of 9 layers of text-based Transformer, 2 layers of panoramic Transformer, a global cross-modal encoder, and a local cross-modal encoder, with the latter two using 4 layers of Transformer each. The other hyperparameters are set as in LXMERT \cite{tan2019lxmert}, with pre-trained LXMERT used for initialization. For feature extraction, we use CLIP-B/16 \cite{radford2021learning} for image feature extraction. Object bounding boxes are provided for the REVERIE dataset, and we similarly use CLIP for feature extraction. Directional features include the sin and cos values of pitch and yaw angles.

During pre-training, we use standard Masked Region Classification (MRC) \cite{lu2019vilbert}, Masked Language Modeling (MLM) \cite{devlin2018bert}, and Single-Step Action Prediction (SAP) \cite{chen2021history} on the R2R dataset. Additional Object Grounding (OG) \cite{lin2021scene} is also used on the REVERIE dataset. To enhance feature representations, we employ EnvEdit \cite{li2022envedit}. The pre-training is conducted on R2R and REVERIE datasets using 5 Tesla V100 GPUs with 16GB memory, with an AdamW optimizer \cite{loshchilov2017decoupled} and a batch size of 6 per GPU, learning rate set to $1 \times 10^{-5}$. The maximum training iterations are 110K for REVERIE and 355K for R2R. In the fine-tuning phase, we use a single NVIDIA 3090 GPU, with a batch size of 6 and a learning rate of $1 \times 10^{-5}$ for both datasets.

\subsection{Comparison with Existing Methods}
Tables \ref{tab:r2r} and \ref{tab:reverie} present a comparison between our method and state-of-the-art methods. On both datasets, our method achieves superior performance in navigation accuracy, instruction following, and object localization, whether in seen or unseen environments.

On the R2R dataset, our method achieves the best results across all metrics for each subset. On the val-seen subset, compared to the baseline DUET, our method reduces NE by 0.18 meters and improves OSR, SR, and SPL by 2\%, 3\%, and 3\%, respectively. On the test set, compared to DUET, NE decreases by 0.56 meters, while OSR, SR, and SPL improve by 5\%, 5\%, and 5\%, respectively. On the REVERIE dataset, compared to the state-of-the-art model VER \cite{liu2024volumetric}, our method outperforms VER in all metrics except for OSR on the val-seen subset, where it is slightly lower by 0.26\%. On the test set, compared to the baseline DUET, our method improves OSR, SR, SPL, RGS, and RGSPL by 6.05\%, 5.17\%, 4.57\%, 3.40\%, and 2.85\%, respectively. Compared to the best model VER, our method improves OSR, SR, SPL, RGS, and RGSPL by 0.74\%, 0.86\%, 1.87\%, 1.40\%, and 1.72\%, respectively.

\subsection{Ablation Studies}

\subsubsection{Ablation of KGL and KGDA}
As shown in Table \ref{tab:all}, we integrated KGL and KGDA into the baseline model separately and conducted ablation experiments on the REVERIE dataset's Val Unseen subset. Adding only KGL and KGDA to the baseline improved the SR and RGS metrics, while the SPL and RGSPL metrics decreased. This is because, without KGL, the instructions input into the model introduce significant data bias, leading the model to explore more paths to reach the target position. Without KGDA, the model cannot dynamically allocate the weight between knowledge and instructions, resulting in equal importance given to both knowledge and instructions. This is clearly unreasonable, as navigation instructions are more important than knowledge, with knowledge merely assisting the navigation.

\subsubsection{Ablation of Dynamic Adjustment in KGDA Module}
We conducted an ablation study of the dynamic adjustment in the KGDA to evaluate the impact of dynamic adjustment on the model. The experimental results are shown in Table \ref{tab:past}. On the Test Unseen subset of the REVERIE dataset, the KGDA with dynamic adjustment improved the OSR, SR, SPL, RGS, and RGSPL metrics by 0.25\%, 0.58\%, 0.73\%, 0.44\%, and 0.64\%, respectively, compared to the KGDA without dynamic adjustment.
\begin{table*}[t]
\centering
\caption{Ablation study results of KGL and KGDA on the REVERIE dataset's Val Unseen}
\label{tab:all}
\begin{tabular}{ccccccc} 
\toprule
Id  & KGL & KGDA & SR$\uparrow$ & SPL$\uparrow$ & RGS$\uparrow$ & RGSPL$\uparrow$ \\
\midrule
1  & $\times$ & $\times$ & 46.98 & 33.73 & 32.15 & 23.03 \\ 
2  & $\checkmark$ & $\times$ & 47.54 & 32.05 & 32.49 & 22.20 \\ 
3  & $\times$ & $\checkmark$ & 47.94 & 32.79 & 32.43 & 22.67 \\ 
4  & $\checkmark$ & $\checkmark$ & \textbf{47.97} & \textbf{34.38} & \textbf{33.29} & \textbf{24.42} \\ 
\bottomrule
\end{tabular}
\end{table*}

\begin{table}[t]
\centering
\caption{Ablation study results of KGL and KGDA on the REVERIE dataset's Test Unseen}
\label{tab:past}
\begin{tabular}{cccccc}
\hline
 & OSR$\uparrow$ & SR$\uparrow$ & SPL$\uparrow$ & RGS$\uparrow$ & RGSPL$\uparrow$ \\ \hline
w/o Dynamic  & 62.71 & 57.10 & 39.90 & 34.84 & 24.27 \\ 
w/ Dynamic  & \textbf{62.96} & \textbf{57.68} & \textbf{40.63} & \textbf{35.28} & \textbf{24.91} \\ \hline
\end{tabular}
\end{table}

\subsection{Qualitative Results}
\begin{figure*}[ht!]
    \centering
    \includegraphics[width=\textwidth]{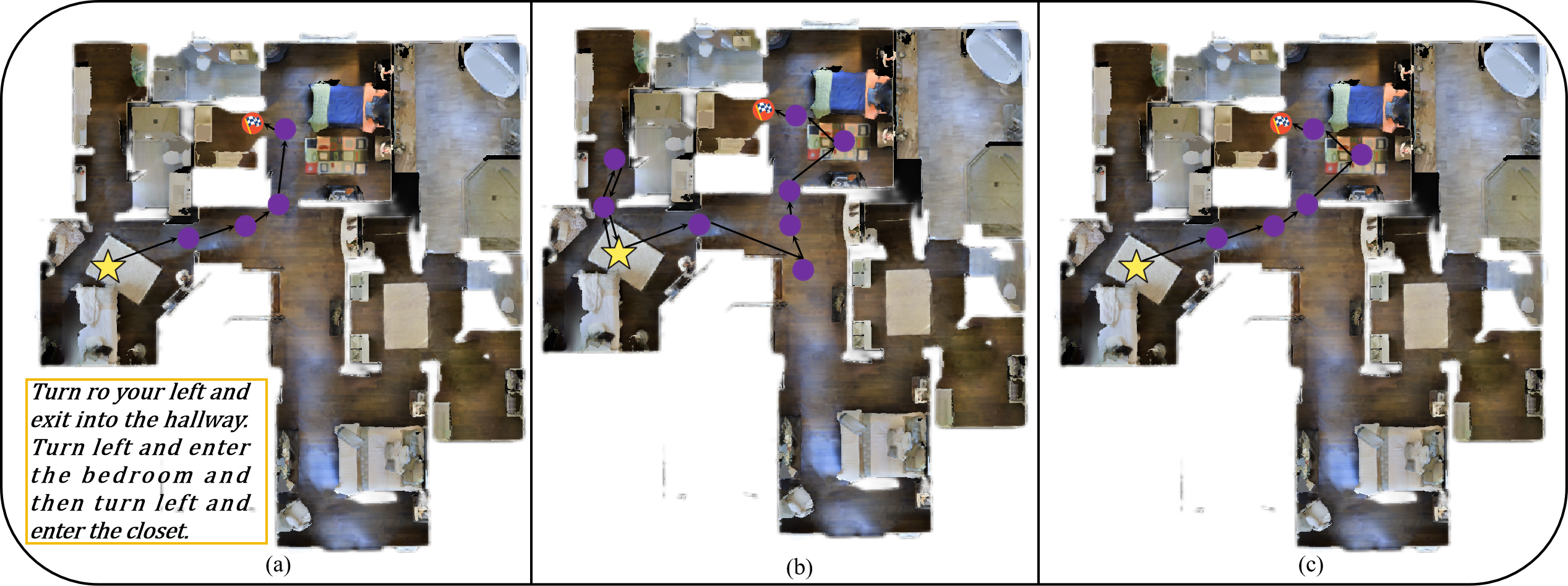}
    \caption{Qualitative results are shown as follows: (a) shows the actual navigation trajectory, requiring 5 movements in total; (b) shows the baseline model's navigation result, which successfully navigates to the target location but requires 11 movements; (c) shows our model's navigation result, which not only successfully navigates to the target location but also requires only 6 movements.}
    \label{fig:guijitu}
\end{figure*}

An example of a navigation trajectory on the unseen validation set of the REVERIE dataset is shown in Figure \ref{fig:guijitu}. In the figure, the sentence in the yellow box is the navigation instruction, the yellow star represents the starting point, the purple circles represent the visited navigation points, and the flag represents the target point. Figure \ref{fig:guijitu}(a) shows the navigation trajectory, which requires 5 movements. Figure \ref{fig:guijitu}(b) shows the navigation result of the baseline model, and Figure \ref{fig:guijitu}(c) shows the navigation result of our model. Compared to the baseline model, our model ensures that the path is as short as possible and correctly reaches the specified location. Our model is better at focusing on landmark information in the instructions and combining knowledge for navigation, allowing it to get the target location more effectively.

\section{Conclusion}
In this work, we propose a novel Landmark-Guided Knowledge (LGK) approach aimed at enhancing reasoning ability and navigation efficiency in vision-and-language navigation tasks. The proposed method effectively mitigates navigation errors caused by the lack of commonsense knowledge in existing approaches, demonstrating particularly strong performance in complex environments. Specifically, LGK first constructs a knowledge base enriched with diverse language descriptions and employs instruction-referenced landmarks to guide the selection and refinement of matched knowledge. It then integrates visual history and language semantics to achieve efficient multimodal information fusion. Experimental results on two mainstream benchmarks, R2R and REVERIE, show that LGK consistently outperforms state-of-the-art methods across multiple key evaluation metrics, validating its effectiveness and generalizability in enhancing the agent’s understanding and execution capabilities. In future work, we plan to explore finer-grained knowledge modeling and more robust reasoning frameworks to further improve the agent’s generalization ability and autonomous decision-making in open-world scenarios.



\section*{Acknowledgements}

This research was financially supported by the National Natural Science Foundation of China (Grant No. 62463029) and the Natural Science Foundation of Xinjiang Uygur Autonomous Region (Grant No. 2015211C288).

%
%
%
%





\bibliography{ref}

\end{document}